\documentclass[letterpaper, 10 pt, conference]{ieeeconf}

\IEEEoverridecommandlockouts                              % This command is only needed if 
\usepackage{lipsum}
\usepackage{wrapfig}
\usepackage{graphicx}
\usepackage{caption}
\usepackage{subcaption}
\usepackage{float}
\usepackage{xcolor}
\usepackage{amsmath}
\usepackage{dblfloatfix}
\usepackage{algorithm}
\usepackage[noend]{algpseudocode}
\usepackage{multicol}
\usepackage{siunitx}
\makeatletter
\def\BState{\State\hskip-\ALG@thistlm}
\title{\LARGE \bf
Autonomous Marine Sampling Enhanced by Strategically Deployed Drifters in Marine Flow Fields
}
\author{Johanna Hansen$^{1}$, Sandeep Manjanna$^{1}$, Alberto Quattrini Li$^{2,3}$, 
Ioannis Rekleitis$^{2}$ and Gregory Dudek$^{1}$% <-this % stops a space
\thanks{$^{1}$Mobile Robotics Lab at the Centre for Intelligent Machines at McGill University, Montreal, QC, Canada {\tt\small jhansen, msandeep,dudek@cim.mcgill.ca}}%
 \thanks{$^{2}$Autonomous Field Robotics Lab, University of South Carolina, Columbia, South Carolina, {\tt\small yiannisr@cse.sc.edu}}
\thanks{$^{3}$Department of Computer Science, Dartmouth College, Hanover, New Hampshire, {\tt\small alberto.quattrini.li@dartmouth.edu}}
\thanks{\textbf{Submitted to the 2018 IEEE/MTS OCEANS Conference. 978-1-4799-7492-4/15/\$31.00~\copyright2018~IEEE}}}
\begin{document}

\maketitle

\thispagestyle{empty}
\pagestyle{empty}

%%%%%%%%%%%%%%%%%%%%%%%%%%%%%%%%%%%%%%%%%%%%%%%%%%%%%%%%%%%%%%%%%%%%%%%%%%%%%%%%$
\begin{abstract}
We present a transportable system for ocean observations in which a small autonomous surface vehicle (ASV) adaptively collects spatially diverse samples with aid from a team of inexpensive, passive floating sensors known as \emph{drifters}. Drifters can provide an increase in spatial coverage at little cost as they  are propelled about the survey area by the ambient flow field instead of with actuators. Our iterative planning approach  demonstrates how we can use the ASV to strategically deploy drifters into points of the flow field for high expected information gain, while also adaptively sampling the space. In this paper, we examine the performance of this heterogeneous sensing system in simulated flow field experiments. 
\end{abstract}

\section{Overview}

% work is to develop a low-cost, non-static sensor system to collect measurements in a flow field with minimal investment by exploiting natural currents to sample trajectories over a survey region. 
In this paper, we present an information-theoretic approach for fast marine surveying with an autonomous surface vehicle (ASV) which strategically augments coverage by strategically deploying floating marine sensors called \emph{drifters}. The drifters are moved about the environment by a flow field consisting of  water current at a fixed depth and provide improved spatial coverage with little added cost to the system in terms of time or expense. Combined data from the ASV and drifters allows the vehicle to adaptively sample information-rich regions and calculate new drifter deployment points. In this paper we explore the utility of this mixed-modality surveying scheme and provide empirical results from ocean flow simulations.

% \begin{figure}[h!]
% \centering
% \includegraphics[width=0.46\textwidth]{{crv_imgs/driftSearchGeneric.png}}
% \caption{A system overview depicting two deployed drifters which relay information about their local environment back to an autonomous boat which is capable of observing the same phenomena. }
% \label{fig:diagram}
% \end{figure}

\begin{figure}
    \centering
   % \hspace{-.5cm}
  \begin{subfigure}[b]{0.33\textwidth}
  \includegraphics[width=\textwidth]{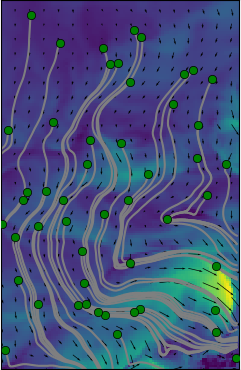}
 % \caption{}
  %\label{fig:ex_flowfield}
 \end{subfigure}   
   \hspace{.0001cm}
   \begin{subfigure}[b]{0.08\textwidth}
 \includegraphics[width=\textwidth,height=9cm]{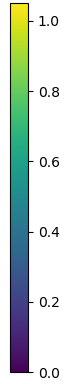}
% \caption{}
 %\label{fig:flowfield_colorbar}
 \end{subfigure}
 \vspace{.2cm}
 \caption{This figure demonstrates the use of the particle simulator, OpenDrift \cite{opendrift}, to determine drifter trajectories in a flow field that was used in our experiments. The drifters depicted in this figure were randomly released at the green points and traveled along the gray tracks over the period of $5$ hours. The flow field direction is described in each grid cell by an arrow and current speed is represented by the colormap in \si[per-mode=symbol]{\m\per\s} according to the color bar. This particular flow field is one of $25$ used for evaluation and serves as the ground truth flow field, $\vec{V}$, for Figures \ref{fig:decision} and \ref{fig:final}.
}
\label{fig:gt_example}
 \vspace{-.5cm}
\end{figure}

Traditionally, environmental surveys of coastal areas required sophisticated (costly) robotic vehicles or significant human effort to effectively model a region. Most autonomous robotic surveying systems employ an exhaustive waypoint-tracking sampling strategy over an unknown survey region which can be tedious and impractical if the survey space is large and/or the phenomenon of interest has only a few regions with important information.  Human-driven sampling systems, however, often make informed  decisions about where to collect data based on their prior knowledge and expertise, preferring to heavily sample unpredictable areas and take fewer measurements from well-modeled areas. Our system seeks to make similar informed sampling choices to reduce the need for the ASV to physically move about a survey region. 

Non-actuated marine drifters like the ones employed in this paper are controllable only at the launch point, but move passively once they are deployed by exploiting the external forces of the local flow field (usually water current). Drifters can be equipped with a variety of sensors for collecting georeferenced data and have been used extensively in oceanography \cite{gpsdrifter, overview_drifters, LumpkinPazos2007}, filling an important niche with their long battery life and low-cost. Although these passive sensors are unable to control their own movement, given knowledge of an ambient flow field and physical characteristics of the device, the trajectory of a drifter can be approximately calculated \cite{RyanDrifter,lumpkinspreading,lumpkinnearsurface} from a given position. In our system, data from deployed sensors (including the boat and deployed drifters) is iteratively collected, assimilated, and used in a particle simulator to estimate the value of sampling a location. %In addition to selectively sampling a phenomena of interest, our system must simultaneously discover the flow field which will propel drifters and use its estimate of the flow field to choose high-value drifter launch locations. 

Our system can be used to collect data for phenomena which are locally observable by the drifters and the ASV with no knowledge of the region before beginning a survey. The ASV and all drifters must be capable of collecting compatibly, transformable samples from $1)$ the specified \emph{phenomena of interest} and $2)$ from the flow field at the same fixed depth. In this paper, we select the \emph{phenomena of interest} to be the flow field itself for the purpose of improving visualization, however, we can easily optimize for a different phenomena such as oxygen, visual observation, or temperature. We leverage predictive modeling to distribute sampling amongst  actuated (relatively costly ASV) and non-actuated (low-cost drifters) sensor platforms so that many inexpensive drifters are exploited, while the ASV intelligently samples regions which are not easily reachable by the non-actuated sensors. We show that this can result in faster surveys in many environments. 

\section{Background}

Although complete sampling of region at a higher than Nyquist frequency is almost always ideal, practical constraints often limit the time or resources that can be used to collect data in real systems. When time allows, often a Boustrophedon \cite{Choset97coveragepath} or \emph{lawnmower} style sampling pattern is used to map an entire survey area. However, in some instances, such as those in which high-variable data clusters in spatial regions, an adaptive sampling approach can yield similar modeling errors with significant reduction in the time/energy resources needed  \cite{data_driven, low2008adaptive,rahimi2005adaptive,singh2007efficient,fiorelli2006multi,plume_tracking}. 

We demonstrate the utility of a non-uniform sampling technique for exploiting expected high-information regions of a survey area using a low-resolution survey by the ASV as a prior \cite{sandeep_iros2017} and samples gathered from randomly deployed drifters as a prior in \cite{crv}. In  \cite{jo_iros18}, we found drifter deployment points which optimized for survey coverage. In this work, we combine our previous work into a comprehensive technique for optimizing modeling error with an ASV which adaptively samples and deploys drifters so as to split the burden of spatial sampling. Others have shown that drifters can be used to exploit the unique nature of flow fields to transport drifters for sampling, search, and exploration \cite{meghjani2016multi, Rekleitis2012b, RyanDrifter,japan_drifter_for_dispersion}. In \cite{coordinated_sampling}, the authors present an approach to multi-day sampling of  time and spatially varying oceanographic phenomena in which drifters inform autonomous underwater vehicles of the movement of features of interest. Like our approach, others have utilized flow fields for tracking features of interest \cite{rssdrift} and for generating informed paths \cite{Kularatne2018a,optimal_trajectory_generation_glider,flow_coverage_drifter}.

\begin{figure}[h!]
\centering
\includegraphics[width=0.48\textwidth]{{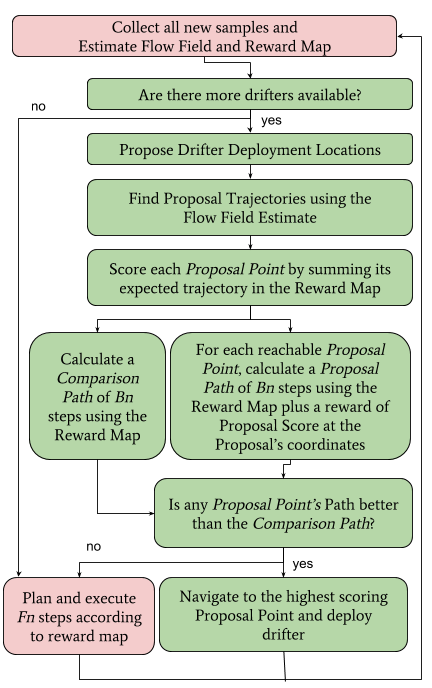}}
\caption{Planning decision process}
\label{fig:flowchart}
\end{figure}

\begin{figure*}[h!]
    \centering
   % \hspace{-.5cm}
  \begin{subfigure}[b]{0.22\textwidth}
  \includegraphics[width=\textwidth,height=6.cm]{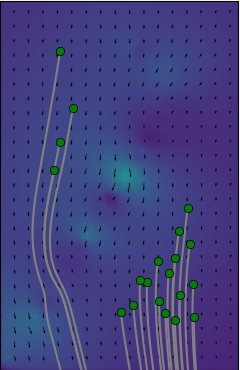}
          \caption{Proposal Points}
          \label{fig:proposals}
 \end{subfigure}   
   \hspace{.0001cm}
  \begin{subfigure}[b]{0.22\textwidth}
  \includegraphics[width=\textwidth,height=6.cm]{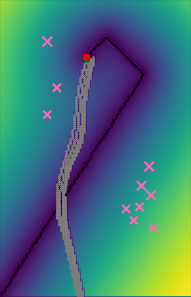}
          \caption{Decision State}
          \label{fig:decision_state}
 \end{subfigure}
      \hspace{.0001cm}
 \begin{subfigure}[b]{0.22\textwidth}
  \includegraphics[width=\textwidth,height=6.cm]{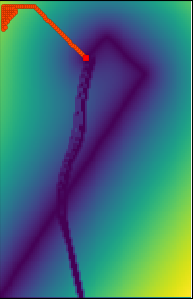}
         \caption{Adaptive Comparison}
         \label{fig:adaptive}
 \end{subfigure}\hspace{.0001cm}
  \begin{subfigure}[b]{0.22\textwidth}
  \includegraphics[width=\textwidth,height=6.cm]{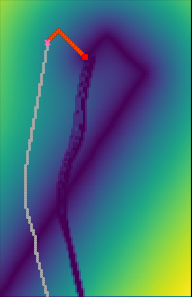}
         \caption{Deploy Comparison}
         \label{fig:deploy}
 \end{subfigure}\hspace{.0001cm}
 \begin{subfigure}[b]{0.04\textwidth}
 \includegraphics[width=\textwidth,height=6.cm]{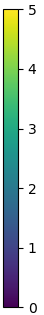}
         \caption{}
         \label{fig:deploy_colorbar}
 \end{subfigure}%\hspace{.0001cm}
 \caption{This figure demonstrates a decision point in which the ASV considers the deployment of a second drifter into the flow field $16$ minutes after starting the experiment. This experiment used $5$ drifters with $F_n=5$, and $B_n=90$.  The current estimate of the flow field is depicted in the background of Figure \ref{fig:proposals} (the true flow field is in Figures \ref{fig:gt_example} and \ref{fig:gt_flowfield}). The process described in Section \ref{sec:proposals} was used to propose the proposal points depicted in Figure \ref{fig:proposals} with their expected trajectories given the current estimate.  The experiment state is depicted in Figure \ref{fig:decision_state} with the ASV's current position show with a red square marker. The ASV has just finished deploying a drifter, which has an expected trajectory plotted in gray. The Proposal Points are plotted as pink $x$ markers with their size correlating to their expected value along their trajectories. The reward map is plotted in the right three plots with their colorbar shown in Figure \ref{fig:deploy_colorbar}. Figures \ref{fig:adaptive} and \ref{fig:deploy} with expected values of $304.64$ and $802.76$ show the two paths considered in red. In this case, the second drifter was deployed. }
  \label{fig:decision}
 \end{figure*}
\section{Method}

Formally, we are considering the problem of physically collecting point measurements over a defined marine region, $S$,  with one ASV and a number, $N_d$ of marine drifters so as to reconstruct the spatially distributed  \emph{phenomena of interest}, $I$,  as accurately as possible with minimal time invested.   The survey space, $S$, is discretized into $m \times n$ square cells of size $r$, where $r$ is related to the sample validity of the sensors used. At each update, we assimilate  samples that have been collected by the ASV and deployed drifters. An update occurs after the ASV has traveled $F_n$ grid cells.

Our drifter deployment approach relies on estimating an observable, but non-uniform flow field, $V$, to evaluate deployment locations. The complete flow field calculation presented in Equation \ref{eq:flow} is simplified in our approach by assuming that $V$ is constant over our observation window and that our observations are taken from a fixed depth. Thus, we can describe  $V$ in Euclidean space  as $\vec{V}=\vec{V}(x,y)$, which is a function of coordinates in space ($x$,$y$) and  the velocity at every point in $\vec{V}$ is defined by components in each coordinate direction as $\vec{V}=u\vec{i}+v\vec{j}$. 

\begin{equation}
\vec{V}=u(x,y,z,t)\vec{i} + v(x,y,z,t)\vec{j} + w(x,y,z,t)\vec{k}
\label{eq:flow}
\end{equation}

With a perfectly known flow field, $\vec{V}$, and an initial location of point particle, it is possible to calculate trajectory of the particle through $\vec{V}$  using the advection equation. However, since at the beginning of the survey, the flow field is largely unknown and our drifters are not point particles, but physical devices with imperfect models, we can only roughly estimate trajectories from a particular deployment point. In order to find deployment locations which will results in high-information gain trajectories, we first estimate the flow field from observed data and then use a computationally tractable process similar to what was first described in \cite{jo_iros18} to propose and evaluate launch points.

\subsection{Modeling the Survey Region}
\label{sec:assimilation}

We employ a Gaussian process (GP) \cite{neal1994priors,gpy2014} for assimilation of $I$ and $V$ data points. Gaussian processes have been widely used for data assimilation and robot-planning \cite{assimilation88,  salmon2006, KAMACHI1995159, tinkaquadratic, no-regret-planning} because of their sample efficiency and uncertainty measure.  We formulate this problem of predicting the unsampled points of $S$ as a Bayesian regression problem. The GPs used in these experiments have  exponential kernels, $K(a,a') = \sigma^2 \exp(\frac{(a - a')^2}{2 l^2})$, with $\sigma=1$, $l=4$. In this notation, $l$ is the \emph{length scale} parameter which regulates how far the GP will extrapolate from an observed data point and $\sigma$ is a scaling factor that determines the average distance the function will be from the mean. The covariance matrix of the GP, $U$, provides us with a measure of uncertainty of the estimate that describes the similarity between every pair of input points $K(a,a')$.  We use a modified version of the uncertainty measure of our estimate $\hat{I}$ of the true $I$ as the \emph{Reward Map}, $R$ which will inform the ASV's adaptive path planning and the selection of drifter deployment points. 

\subsection{Adaptive Path Planning}
\label{sec:planning}
To drive exploration and mapping into areas of expected high information gain we use the uncertainty estimate from the assimilation step, $R$, as a reward function for finding the value of sampling each cell in $S$. We utilize Value Iteration, an approach for finding optimal policies as defined by the Bellman equation in a Markov Decision Process  \cite{bellman1957dynamic}.  The value of sampling each cell from a given state is described by $V^*$ (Equation \ref{eq:value_func}). Once $V^*$ is found, the optimal policy $\pi^*$ from a given state ($s$) can be found by taking the action $a$ with the largest value. In our experiments, we set the discount factor, $\gamma=0.95$. 
\begin{equation}
\label{eq:value_func}
  V^*(s) = \max_a\left(R(s,a) + \gamma \sum_{s' \in S} p(s'|s,a) V^*(s')\right),
\end{equation}

\begin{equation}
\label{eq:policy}
  \pi^*(s) = arg \max_a\left(R(s,a) + \gamma \sum_{s' \in S} p(s'|s,a) V^*(s')\right)
\end{equation}
Ideally, for full Markovian guarantees, after each sample is collected, data assimilation would be recomputed and a new reward map calculated, however, this is computationally intractable. Empirically, we have found that performing a data assimilation update every $F_n=15$ steps to work well in practice. We negate the reward map at the coordinates of each step as it is planned with Value Iteration as an approximation to the full recomputation.  For full details regarding this sequential decision making process, refer to our previous work \cite{sandeep_iros2017}.
 %finite-horizon, partially-observed sequential decision processes.

\begin{figure*}

    \centering
   % \hspace{-.5cm}
  \begin{subfigure}[b]{0.22\textwidth}
\includegraphics[width=\textwidth,height=6.cm]{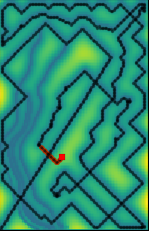}
\caption{Finished Path}
\label{fig:boat_path}
\end{subfigure}   \hspace{.0001cm}
\begin{subfigure}[b]{0.22\textwidth}
\includegraphics[width=\textwidth,height=6.cm]{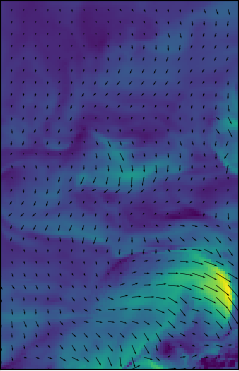}
\caption{True Flowfield Estimate}
   \label{fig:gt_flowfield}
\end{subfigure} \hspace{.0001cm}
\begin{subfigure}[b]{0.22\textwidth}
\includegraphics[width=\textwidth,height=6.cm]{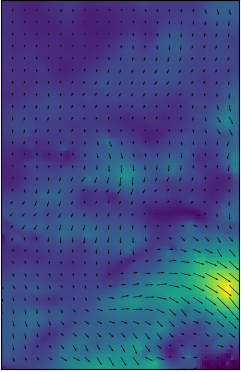}
\caption{Final Flowfield Estimate}
\label{fig:final_flowfield}
\end{subfigure} \hspace{.0001cm}
\begin{subfigure}[b]{0.22\textwidth}
\includegraphics[width=\textwidth,height=6.cm]{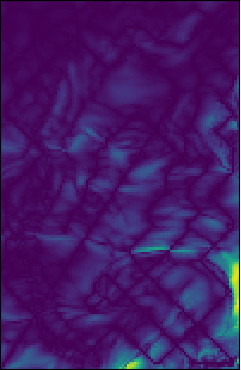}
\caption{Final RMSE}
\label{fig:final_mse}
\end{subfigure} \hspace{.0001cm}
 \begin{subfigure}[b]{0.045\textwidth}
\includegraphics[width=\textwidth,height=6.cm]{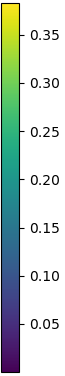}
\caption{}
\label{fig:final_mse_colorbar}
\end{subfigure}
    
\caption{Figure \ref{fig:boat_path} shows the full boat path (black points) at the end of an experiment (last position shown in red) and all sampled positions (violet). The background is the current \emph{Reward Map}, $R$ which references the colorbar shown in Figure \ref{fig:deploy_colorbar}. This experiment used $5$ drifters with $F_n=5$, and $B_n=90$.  Figure \ref{fig:gt_flowfield} is the true flow field and Figure \ref{fig:final_flowfield} is what the experiment estimated the flow field to be at the end of the experiment. The backgrounds Figures \ref{fig:gt_flowfield} and \ref{fig:final_flowfield} refer to speed in \si[per-mode=symbol]{\m\per\s} in the colorbar seen in Figure \ref{fig:gt_example}. The rightmost image, Figure \ref{fig:final_mse}, shows the Root Mean Square Error (RMSE) in \si[per-mode=symbol]{\m\per\s} of Figure \ref{fig:final_flowfield} with respect to the ground truth. The colorbar in Figure \ref{fig:final_mse_colorbar} provides the metric for Figure \ref{fig:final_mse}. }
\label{fig:final}  
\end{figure*}
 
\subsection{Strategic Drifter Deployment}
\label{sec:proposals}
We use a particle trajectory modeling framework, OpenDrift \cite{opendrift}, to estimate likely paths of drifters with given starting points under our estimate of the flow field $\hat{V}$. We model trajectories twice in each update, first to predict hypothetical trajectories for proposed deployment locations and again to estimate the future paths of drifters which are already deployed so as to account for their future in the Reward Map, $R$.

For each deployed drifter, we seed $n_f$ points ($n_f=5$ in these experiments) in a  \SI{2}{\m} radius around the last known location and use OpenDrift to find possible future trajectories so we can update $R$ to reflect the best estimate of where the drifters will travel. For each deployed drifter,
$\frac{1}{n_f}$ is subtracted from  $R$ at each point in the estimated future trajectories and is used in path planning and finding proposal points. The expected future trajectory of a newly deployed drifter is shown in gray in Figure \ref{fig:decision_state} and is reflected in the $R$ which corresponds to the background of Figures \ref{fig:adaptive} and \ref{fig:deploy}. 

We assume that all $N_d$ drifters used in the experiment are carried on the ASV and can be autonomously deployed in $D_p$ seconds (where $D_p=10$ in our experiments).  The deployment of a drifter takes time and energy to travel to an appealing launch location, but thereafter, the data collected from the drifter comes at no cost to the ASV. 

It is computationally expensive to predict drifter trajectories, so we use a method of reducing the number of points to evaluate as originally described in previous work, \cite{jo_iros18}.  The multistage proposal process was inspired by modern object recognition systems such as \cite{faster-r-cnn}. Spatially diverse points are generated by performing rejection sampling over likelihood matrix of size $S$ consisting of a normalized combination of $R$, Gaussian noise, and a safety buffer. This likelihood matrix makes proposal points in areas that are high-uncertain locations more likely.  The safety buffer ensures that no points will be proposed near the edge of the survey space. In this paper, we sample $1000$ unique points with rejection sampling over the likelihood matrix and then prune these points using non-maximum suppression (NMS) until we have a chosen a few points to evaluate with the particle simulator ($20$ in this paper). NMS greedily selecting high-value proposals while deleting nearby proposals which cover the same area \cite{nms} so as to achieve spatially diverse points.

Trajectories are found for the top proposal points by simulating paths from each of the points for the expected battery life of the drifters ($4$ hours in these experiments) with our best estimate of the flow field, $\hat{V}$ (shown in Figure \ref{fig:proposals}). The resulting trajectories are scored by summing the expected sampling points in $R$. The highest scoring proposal points (the top $10$ of the $20$ evaluated in the experiments presented here) are then passed to the ASV decision process for consideration (proposal point score is shown by their respective size in the pink $x$s in Figure \ref{fig:decision_state}). 

\section{Decision State}
\label{sec:decision_state}
While the ASV still has undeployed drifters, it must decide at each update step whether to deploy a drifter at a proposal point or adaptively sample the area. Our system provides a variable from which to control this decision point, $B_n$. At each of these decision points, the ASV plans a \emph{comparison path} using Value Iteration of $B_n$ steps and finds the total score that would be achieved with the $R$ through those points. Alternatively, for each proposal point, the ASV calculates a path of maximum length $B_n$ which travels through the point and finds the score of this path plus the expected value of the proposal point which was calculated in the previous step. This can be thought of as planning a non-optimal path to the proposal point, but with a bonus of earning the full trajectory of the drifter for free. If any of the proposal planned paths scores higher than the comparison path, then that path is executed up to the proposal point and the drifter is deployed. 

If the comparison path scores better than any of the proposal paths or if there are no drifters left to deploy, then $F_n$ steps are executed. At this point, data assimilation is restarted and the process repeats itself, as seen in Figure \ref{fig:flowchart}. 

\begin{figure}[h!]
\centering
\includegraphics[width=0.48\textwidth]{{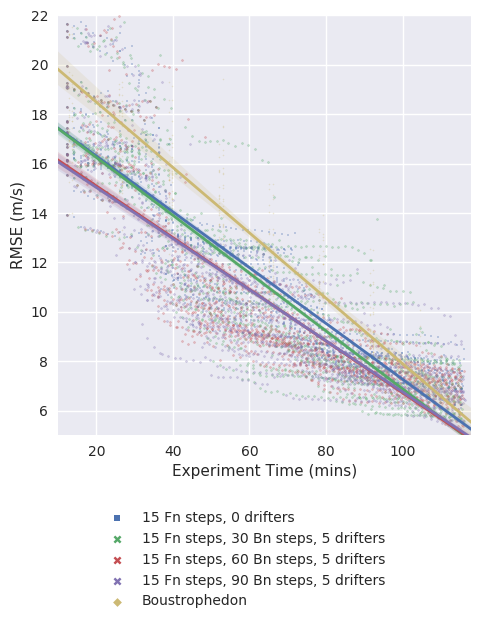}}
\caption{This plot shows the RMSE at each point in which data was assimilated over all $25$ test maps. The lines indicate the trend found with a first-order, bootstrapped regressor ($n=1000$) with shading showing the $95\%$ confidence interval.  This figure demonstrates the advantage of using a tuned adaptive drifter system to effectively model a region quickly. Given enough survey time, the different schemes converge. }
\label{fig:mse_15}
\end{figure}

\begin{figure}[h!]
\centering
\includegraphics[width=0.48\textwidth]{{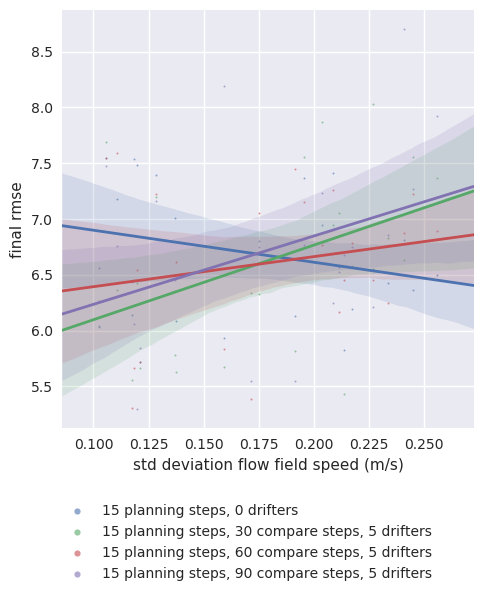}}
\caption{This figure shows resulting final RMSE after the ASV sampled for $2$ hours. Drifter data was integrated until the drifter exited the survey area or expired. Our approach outperforms the adaptive planning approach when the flow field is more predictable. However, when the flow field is less predictable, drifters are sometimes placed in sub-optimal deployment points due to a poor estimate of the true data. }
\label{fig:std_15}
\end{figure}

\section{Experiments}

In order to obtain quantitative results under ground truth, we evaluate our approach using $25$ archival flow fields from a Regional Ocean Modeling Systems (ROMS) dataset with simulated sensor and vehicle placements. This data set consists of an array of  $100\times155$  measurements of in-situ ocean current.  We rescaled the original dataset from \SI[product-units = single]{800 x 800}{\metre} grid cell size to an interval of \SI[product-units = single]{5 x 5}{\metre} in order to make the region feasibly traversable for a typical battery-powered ASV. As a result, we compared our results over $40$ different ocean flows over a region of size \SI[product-units = single]{500 x 775}{\metre}. 

Our simulated ASV kept an average speed of \SI[per-mode=symbol]{1}{\m\per\s} and is capable of choosing $1$ of $8$ actions corresponding to movement into adjacent cells at each time step.  We assume that neither the flow field nor the drifter load have any effect on the speed of the ASV. Our simulation allows the ASV to operate for $2$ simulated hours after the experiment begins. 
\begin{figure*}[htb!]
\centering
\begin{subfigure}[b]{0.30\textwidth}
        \includegraphics[width=\textwidth,height=3.6cm]{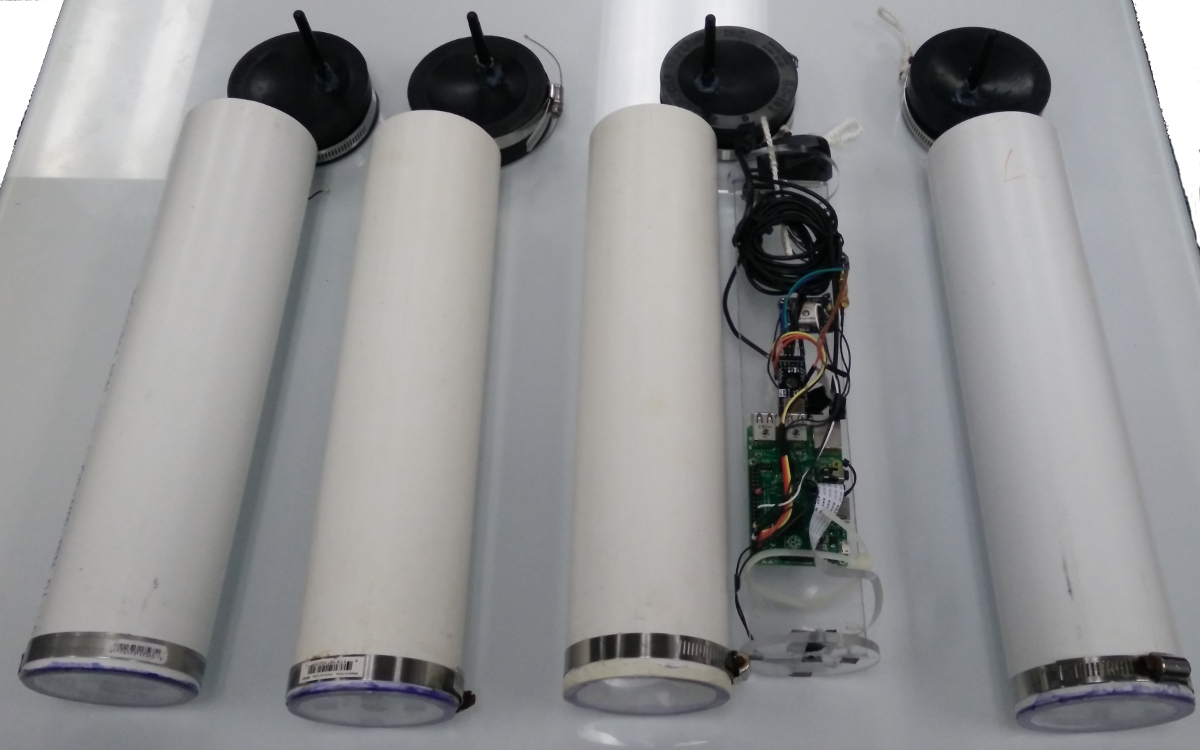}
        \caption{ocean drifter}
        \label{fig:drifter}
    \end{subfigure}\hspace{.0001cm}
       \begin{subfigure}[b]{0.30\textwidth}
        \includegraphics[width=\textwidth,height=3.6cm]{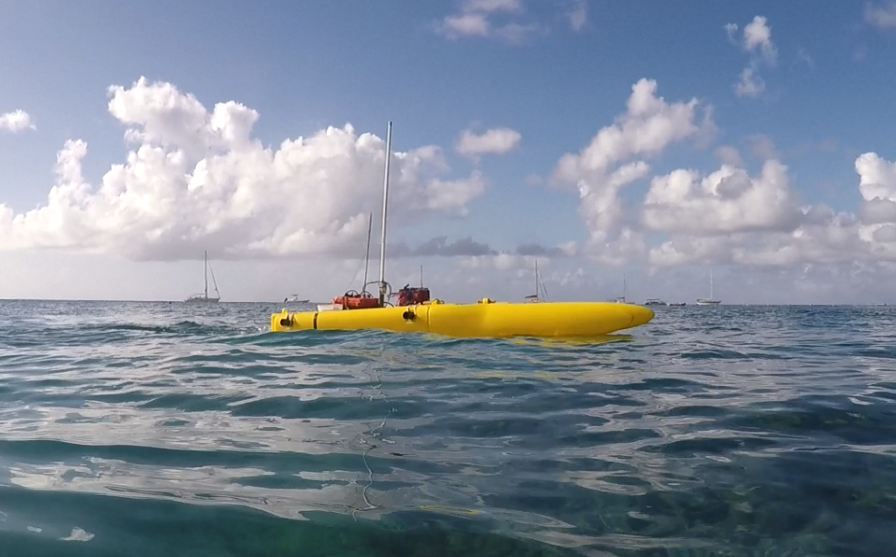}
        \caption{autonomous surface vehicle}
        \label{fig:boat}
    \end{subfigure}\hspace{.0001cm}
    \begin{subfigure}[b]{0.30\textwidth}
        \includegraphics[width=\textwidth,height=3.6cm]{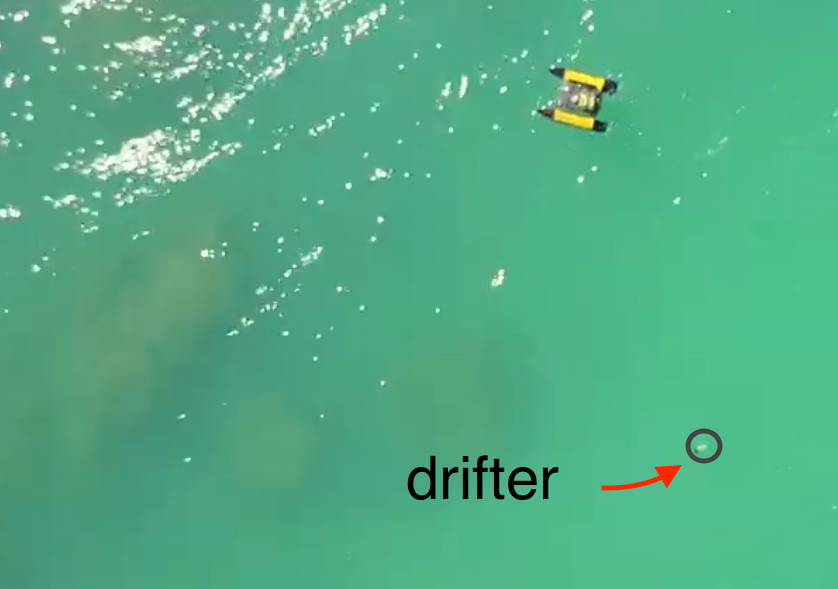}
        \caption{overhead of deployment}
        \label{fig:overhead}
    \end{subfigure}\hspace{.0001cm}
    \caption{Hardware description for field experiments.}
    \label{fig:setup}
    \vspace{-1em}
   \label{fig:hardware}
\end{figure*}

Each released drifter will take a sample every $5$ seconds until it exits $S$ or its time limit expires. In the simulations depicted  here, the drifters collect data for $4$ hours after their initial release. The entire experiment terminates when all deployed drifters have left $S$ or have expired, though the boat stops sampling after it reaches its time limit. 

Each adaptive experiment starts the same way, with the ASV at coordinate $(0,0)$ and driving $75\%$ of the way to the diagonal corner of the survey area. After reaching this point, the first update is run, complete with data assimilation, deployment proposals, path comparisons, and finally driving the next path. 

In Figure \ref{fig:mse_15} we show results from all $25$ simulated flow fields.  Our system has an initial advantage over the ASV-only deployments, as it is able to achieve a more comprehensive model faster.  As expected, however, after appropriate survey time, the ASV-only experiments sufficiently cover the region  in a more complete manner. The selection of $B_n$ for a survey will determine how selective the ASV is when deploying drifters. We see in Figure \ref{fig:std_15} that surveys in flow fields with more variability perform better without drifters of with large $B_n$. In the future, we hope to learn and adapt this parameter during surveys. 

We have also conducted preliminary field experiments (see Figure \ref{fig:hardware}) which demonstrate the proof of concept of our system. We were able to successfully communicate with distributed sensors over WiFi and assimilate flow data collected from GPS measurements. At the time of the field experiments, we lacked an automated deployment mechanism.  

It is also important to consider the added complexity of adding drifters to a survey team. Each additional sensor adds more hardware that must be maintained and repaired. In addition, if drifters are to be reused, they must be physically retrieved at the end of the experiment. Although they will typically have long battery life, drifters almost always cover a region more slowly than actuated vehicles. 

% TODO - it would be nice to plot mse for just the example
% then perhaps give several examples - where ours performed best vs worse. 
% should discuss how to choose B_n and the results
% discuss - drifters tend to be slow so difficult to  use
% discuss fig 6
% plot results from success and failure case?
% calculate rmse [0] for each map and place in for figure 5. discuss battery
% figure 6 - should discuss
% future work?

\section{Conclusion}
We investigate relationships between survey time, energy, and modeling error, and present a tunable algorithm for selectively choosing when to add additional drifters to the survey. We also show when drifters are not useful such as when there is an unstable or low-velocity flow field or when survey time is not limited.

Practical environmental surveys require trade-offs between cost, mobility, and spatial or temporal resolution. By exploiting the efficiencies of both active and passive sensor platforms, we are able to effectively observe environmental phenomena. We show that the proposed heterogeneous system selectively samples these environments to achieve faster modeling results in many scenarios. 

%\begin{thebibliography}{99}

\bibliographystyle{IEEEtran}
\bibliography{IEEEabrv,bib}

\end{document}